\documentclass{article} 
\usepackage{iclr_set,times}

\usepackage{amsmath,amsfonts,bm}

\def\eqref#1{equation~\ref{#1}}

\def\1{\bm{1}}

\DeclareMathAlphabet{\mathsfit}{\encodingdefault}{\sfdefault}{m}{sl}
\SetMathAlphabet{\mathsfit}{bold}{\encodingdefault}{\sfdefault}{bx}{n}

\usepackage{hyperref}
\usepackage{url}
\usepackage{graphicx}
\usepackage{booktabs} 
\usepackage{multirow} 
\usepackage{colortbl}
\usepackage[linesnumbered,ruled,vlined]{algorithm2e}

\SetCommentSty{mycommfont}

\definecolor{g1}{HTML}{b3e2cd}
\definecolor{r1}{HTML}{fdcdac}
\definecolor{w1}{HTML}{cbd5e8}
\definecolor{b1}{HTML}{fff7bc}
\usepackage{wrapfig}

\newcommand{\centerone}[2]{\multicolumn{1}{>{\columncolor{#1}}c}{#2}}
\newcommand{\centeroneg}[1]{\centerone{g1}{#1}}
\newcommand{\centeroner}[1]{\centerone{r1}{#1}}

\newcommand{\centeroneb}[1]{\centerone{b1}{#1}}

\title{TrustScore: Reference-Free Evaluation of LLM Response Trustworthiness}

\author{Danna Zheng$^{1}$, Danyang Liu$^{1}$, Mirella Lapata$^{1}$, Jeff Z. Pan$^{1, 2}$ \\
            $^{1}$ School of Informatics, University of Edinburgh, UK\\
            $^{2}$ Huawei  Edinburgh Research Centre, CSI, UK\\
        \texttt{\{dzheng, danyang.liu\}@ed.ac.uk}, \; 
        \texttt{mlap@inf.ed.ac.uk}\\
        \texttt{http://knowledge-representation.org/j.z.pan/}
}

\iclrfinalcopy 
\begin{document}

\maketitle

\begin{abstract}
Large Language Models (LLMs) have demonstrated impressive capabilities across various domains, prompting a surge in their practical applications. However, concerns have arisen regarding the trustworthiness of LLMs' outputs, particularly in closed-book question-answering tasks, where non-experts may struggle to identify inaccuracies due to the absence of contextual or ground truth information. 
This paper introduces \texttt{TrustScore}, a framework based on the concept of Behavioral Consistency, which evaluates whether an LLM's response aligns with its intrinsic knowledge.
Additionally, \texttt{TrustScore} can seamlessly integrate with fact-checking methods, which assesses alignment with external knowledge sources.
The experimental results show that \texttt{TrustScore} achieves strong correlations with human judgments, surpassing existing reference-free metrics, and achieving results on par with reference-based metrics. 
\end{abstract}

\section{Introduction}
Large-scale language models (LLMs) have recently been in the spotlight due to their impressive performance in various NLP tasks, sparking enthusiasm for potential applications~\citep{kaddour2023challenges,bubeck2023sparks}.
However, a notable concern has emerged regarding the ability of LLMs to generate plausible yet incorrect responses~\citep{tam2022evaluating,liu2023trustworthy,devaraj2022evaluating}, particularly challenging for users without specialized expertise.
Consequently, users are often advised to employ LLMs in scenarios where they can confidently assess the information provided.

This concern is particularly salient in closed-book question-answering tasks~\citep{roberts2020much,su2023context}, where the absence of contextual cues complicates the evaluation process. 
Unlike tasks such as reading comprehension ~\citep{yao2023korc}, summarization~\citep{aharoni2023multilingual}, and natural language inference~\citep{chen2023menli}, where answers can be verified against provided context, assessing the trustworthiness of an LLM's response in a closed-book setting poses a greater challenge. 
 In such settings, LLMs rely solely on their parametric knowledge~\citep{PRKS+2023} to generate answers, making it difficult to determine whether these responses truly align with their underlying knowledge due to the black-box nature of LLMs. Traditional approaches like fact-checking, which rely on evidence from external knowledge bases~\citep{PVGW2017,Pan2017b}, may be hindered by inaccessible or insufficient information, rendering them ineffectively.

To address these challenges, this paper introduces \texttt{TrustScore}, a framework centered around the concept of \textit{Behavioral Consistency}, which evaluates whether an LLM's response aligns with its parametric knowledge by assessing whether the LLM makes consitent choices among its responses and other distractors.
Behavioral consistency serves as a necessary condition for ensuring an LLM's responses are in alignment with its parametric knowledge, enhancing the likelihood of trustworthiness.
Furthermore, \texttt{TrustScore} can seamlessly integrate with fact-checking methods, offering a more holistic evaluation of an LLM's response. 
We evaluate the effectiveness of \texttt{TrustScore} on FLAN-T5 \citep{chung2022scaling}, LLaMA \citep{touvron2023LLaMA}, and \href{https://openai.com/blog/gpt-3-5-turbo-fine-tuning-and-api-updates}{GPT-3.5}, showing strong correlation with human judgments and outperforming existing reference-free~\footnote{Reference in this paper refers to gold answers.} metrics while achieving performance comparable to reference-based metrics. 


\section{Proposed Method}
Given an LLM denoted as $M$, a context-free question represented as $q$, and the corresponding response generated by LLM $M$ for question $q$ denoted as $a$, the objective of \texttt{TrustScore} is to provide a relative score to show to what extent can the response be trusted. The resulting \texttt{TrustScore} falls within the range [0, 1]. A higher score means the response is more likely to be trusted.

\begin{wrapfigure}{r}{0.45
\textwidth}
\vspace{-1.3em}
\begin{center}
\includegraphics[width=\linewidth]{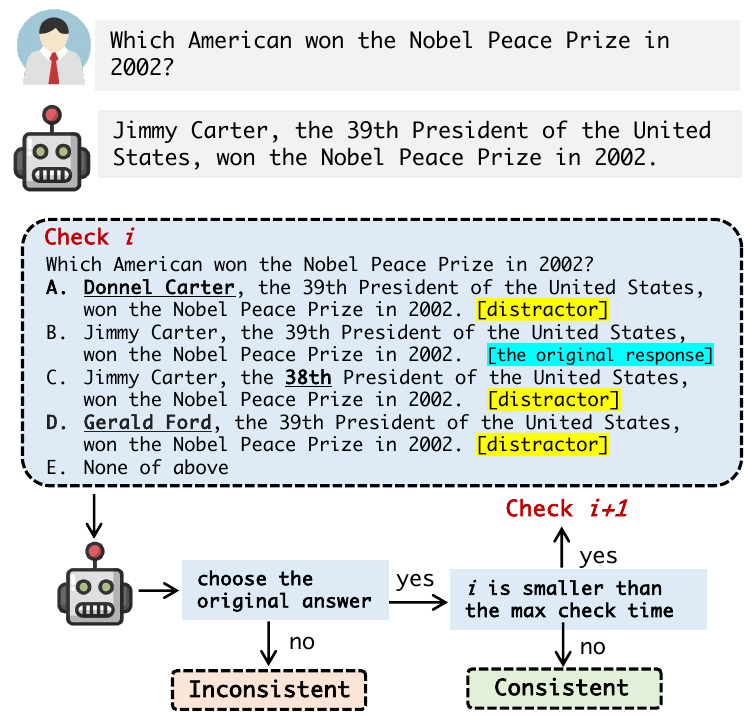}
\end{center}
\vspace{-1.3em}
\caption{An illustration of Behavioral Consistency Evaluator (Trust$_{BC}$). LLM responses pass through Trust$_{BC}$ to determine if the model selects its original answer consistently across multiple iterations.}
\vspace{-2.3em}
\label{fig:trust_frame}
\end{wrapfigure}

The key component of \texttt{TrustScore} is a Behavioral Consistency (Trust$_{BC}$) evaluator as illustrated in Figure~\ref{fig:trust_frame}, which assesses whether the responses provided by the LLM are consistent with its intrinsic parametric knowledge. This evaluator is designed to function independently, particularly useful in situations where external knowledge bases are not accessible, or there is not enough information available.
Moreover, Trust$_{BC}$ can integrate with a Fact-Checking (Trust$_{FC}$) module to further assess trustworthiness using an external knowledge base when one is available. This dual approach allows for a comprehensive assessment of an LLM's response, ensuring it is evaluated both for internal consistency and external factual consistency.

\subsection{Trust$_{BC}$ Stand-Alone}

The Behavior Consistency Evaluator aims to determine whether the LLM $M$ consistently selects response $r$ when presented with distractors $R = \{r_1, r_2, r_3, ...\}$. 
We set a maximum check limit of $n$ times, where LLM $M$ is tasked with choosing the correct option from five choices (including response $r$, three random distracting options, and `None of the above') in each attempt. 
Note that the distracting options are different across different attempts.
If LLM $M$ does not choose response $r$ in any of the attempts, it implies that it does not consistently provide the response to question $q$ in line with its parametric knowledge. 
Ideally, if we conduct an infinite number of checks and LLM $M$ consistently selects response $r$, it implies that response $r$ is genuinely rooted in the LLM's underlying knowledge. 
The behavior check process terminates when inconsistent behavior occurs (LLM $M$ does not select response $r$) or when it reaches the maximum check limit $n$. 
If inconsistent behavior occurs, \texttt{Trust$_{BC}$} return 0; otherwise, if inconsistent behavior does not occur up to the maximum check limit $n$, \texttt{Trust$_{BC}$} return 1.

To generate high-quality distractors, we propose a priority-based substitution algorithm:

\textbf{Words necessitating substitution are prioritized based on their informativeness.} (1) Words exclusively present in response $r$ are given higher priority than those occurring in both response $r$ and question $q$; (2) Entities take precedence over other words; (3) Nouns and numbers are prioritized over verbs and adjectives, which, in turn, take precedence over words of other parts of speech.

\textbf{Substitute words are prioritized based on their sources.} (1) DBpedia entities filtered by types are top priority; (2) words filtered by embedding similarity have the second-highest priority; (3) words randomly sampled from the vocabulary filtered by POS tags are the lowest priority.

Details of the priority-based substitution algorithm are in Appendix~\ref{app:sub_alg}. Examples of the generated multi-choice questions are provided in~\ref{app:mc-q}. 

\subsection{Integration of Trust$_{BC}$ and Trust$_{FC}$}
When external knowledge bases are available, Trust$_{BC}$ can collaborate with Trust$_{FC}$ for a thorough evaluation of responses. 
The role of Trust$_{FC}$ is to verify if a response $r$ to a question $q$ aligns with facts from external knowledge bases. 
This project does not seek to pioneer new fact-checking methods. We adopt a widely accepted approach of extracting relevant evidence from knowledge bases and assessing its support for the response $r$ through an entailment model. 
Trust$_{OV}$ combines scores from Trust$_{BC}$ and Trust$_{FC}$ based on criteria detailed in Appendix~\ref{app:score}. This integration is guided by several principles: (1) Responses supported by external knowledge are deemed more trustworthy than those contradicted by it. (2) A response insisted upon by LLM is more trustworthy than the response not insisted upon by LLM. (3) Factual consistency is prioritized over behavioral consistency.



\section{Experiment Setup}
\paragraph{Data Collection}
We have created a composite dataset, MixedQA, for the purpose of analyzing the correlation between \texttt{TrustScore} and human evaluations.
It consists of 1,000 open-ended questions randomly selected from the test sets (or development sets when test sets are unavailable) of five datasets: Natural Questions \citep{kwiatkowski2019natural}, WebQuestions \citep{berant2013semantic}, TriviaQA \citep{joshi2017triviaqa}, HotpotQA \citep{yang2018hotpotqa}, and PopQA \citep{mallen2022not}.
We ignore the support context in these datasets to fit the closed-book setting.
The detailed selection criteria are in the Appendix~\ref{app: data selection}. The dataset and code are public available~\footnote{https://github.com/dannalily/TrustScore}.

\paragraph{LLMs}
We test our \texttt{TrustScore} on the answer generated by FLAN-T5-XXL, LLaMA-7B, and GPT-3.5-Turbo. LLMs perform the question-answering task in a few-shot setting. To fully exploit \texttt{TrustScore}, we prompt LLM to output response in long sentence instead of short phrase. The implementation detail including the prompts and hyper-prameters are provided in Appendix~\ref{app:implementation}.

\paragraph{Baseline Metrics}
We employ the following baselines: reference-based metrics, reference-free metrics with and without evidence retrieval.
Reference-based metrics include popular lexical matches like Exact Match, BLEU \citep{papineni2002bleu}, ROUGE-L \citep{lin2004rouge}, METEOR \citep{banerjee2005meteor}, and pretrained evaluators such as BERTScore \citep{zhang2019bertscore}, BARTScore \citep{yuan2021bartscore}, and GPTScore\footnote{GPTScore includes several backbones, and we chose FLAN-T5-base in this work.} \citep{fu2023gptscore}.
Reference-free metrics include BARTScore and GPTScore, both with and without evidence retrieval, for a fair comparison with our \texttt{TrustScore}.

\paragraph{Evaluation}
\begin{wraptable}{r}{0.48\textwidth}
\vspace{-2.5 em}
\caption{Pearson’s $r$ correlation for baseline metrics and \texttt{TrustScore} against human judgments in MixedQA. \textit{NAN} indicate uncomputable correlations when all exact match scores are 0. Unless underlined, all reported correlations are statistically significant at $p<0.01$.}\label{tab:trust-correlation}
\centering
     \resizebox{\linewidth}{!}{
        \begin{tabular}{l|c|c|c}
            \toprule
            \multicolumn{1}{c}{\multirow{1}{*}{\textbf{Metrics}}}& \centeroneg{\textbf{FLAN-T5}}& \centeroneb{\textbf{LLaMA}}& \centeroner{\textbf{GPT-3.5}} \\
            \midrule
            \multicolumn{4}{c}{\textit{Reference-based Metrics}} \\
            \midrule
                Exact Match & 0.635 & \textit{NAN} & \textit{NAN} \\
                BLEU-1 & \centeroneg{0.655} & 0.435 & 0.289 \\
                BLEU-4 & 0.536 & 0.376 & 0.228 \\
                ROUGE-L & 0.607 & 0.533 & 0.382 \\
                METEOR & 0.603 & 0.536 & 0.106 \\
                BERTScore & 0.573 & 0.266 & 0.087 \\
                BARTScore-ref & 0.570 & 0.474 & 0.368 \\
                GPTScore-ref & 0.610 & \centeroneb{0.575} & \centeroner{0.453} \\
            \midrule
            \multicolumn{4}{c}{\textit{Reference-free Metrics (w/o Evidence Retrieval)}} \\
            \midrule
                BARTScore-src & -0.075 & -0.099 & -0.151 \\
                GPTScore-src & \underline{-0.055} & \underline{-0.081} & -0.139 \\
                Trust$_{BC}$ (Ours) & \centeroneg{0.313} & \centeroneb{0.133} & \centeroner{0.353} \\
            \midrule
            \multicolumn{4}{c}{\textit{Reference-free Metrics (w/ Evidence Retrieval)}} \\
            \midrule
                BARTScore-evid & 0.070 & 0.091 & -0.013 \\
                GPTScore-evid &	0.075 & 0.186 & 0.024 \\
                Trust$_{FC}$ &	0.593 & 0.533 & 0.449 \\
                Trust$_{OV}$ (Ours) & \centeroneg{0.613} & \centeroneb{0.539} & \centeroner{0.484} \\
            \bottomrule
        \end{tabular}
        }
        \vspace{-4.5em}
\end{wraptable}

To validate the effectiveness of \texttt{TrustScore}, we calculate the correlation between \texttt{TrustScore} and human judgments on the MixedQA dataset.
Humans annotators applied binary judgments to categorize the 3,000 answers generated by LLMs as "correct" or "incorrect."
We initiated this with a pilot annotation of fifty examples, achieving a strong agreement score of 0.89.
Detailed information about the annotation process is provided in Appendix~\ref{app:ann}.

\section{Results}
Table~\ref{tab:trust-correlation} shows Pearson's $r$ correlations between various metric scores and human judgments for answers produced by LLMs.

\paragraph{Behavioral Consistency Analysis}

In experiments where retrieved evidence is unavailable, our Trust$_{BC}$ demonstrates the highest correlation with human judgments, outperforming other reference-free metrics.
Remarkably, Trust$_{BC}$ also outperforms some reference-based metrics when evaluating responses from GPT-3.5.
This is particularly significant given that, in many real-world scenarios, relevant evidence is unavailable or insufficient, making reference-based or retrieval-based metrics inapplicable. In such cases, our Trust$_{BC}$ can effectively assess response trustworthiness. Comparing the Trust$_{BC}$ scores across different LLMs, we observe that the Trust$_{BC}$ score for LLaMA is lower than those for FLAN-T5 and GPT3.5.
This difference could be attributed to LLaMA being a base model, while FLAN-T5 is fine-tuned with instructions and GPT-3.5 is fine-tuned with both instructions and human feedback. These fine-tuning processes may enhance the behavioral consistency of the latter models.

\paragraph{Integration of Trust$_{BC}$ and Trust$_{FC}$} 
In experiments where retrieved evidence is available, Trust$_{OV}$ demonstrates human correlations that surpass those of existing reference-free metrics and even outperform most reference-based metrics.
We note that the Trust$_{OV}$ score consistently outperforms Trust$_{FC}$ across various LLMs. This observation substantiates our assertion that Trust$_{BC}$ is not only an effective reference-free metric on its own but can also complement Trust$_{FC}$ to achieve better performance in a reference-based setting.

\paragraph{Robustness to Diverse Answers}
From Table~\ref{tab:trust-correlation}, we observe that traditional lexical overlapping-based metrics (e.g., Exact Match, BLEU, and ROUGE, etc.) achieve higher correlation scores for responses generated by FLAN-T5, but lower scores for answers generated by GPT3.5.
This discrepancy can be attributed to FLAN-T5's tendency to produce concise answers.
Some of these answers match reference answers exactly, making them easier to evaluate using reference-based metrics.
Conversely, GPT-3.5 tends to generate longer and more diverse answers, which may deviate from reference answers, making lexical-based metrics less effective. In contrast to these lexical overlapping-based metrics, \texttt{TrustScore} excels in evaluating GPT-3.5-generated responses and also achieves comparable results for FLAN-T5-generated responses. This highlights its robustness across different types of answers.

\section{Related Work}

\paragraph{QA Evaluation Metrics} Traditional metrics such as Exact Match, F1, BLEU \citep{papineni2002bleu}, Rouge \citep{lin2004rouge}, and METEOR \citep{banerjee2005meteor}, rely on lexical matching, often overlooking semantic content. BERTScore \citep{zhang2019bertscore}, BARTScore \citep{yuan2021bartscore} and GPTScore\citep{fu2023gptscore} addresses this limitation by leveraging contextualized embeddings to capture semantic similarity. However, these methods are predominantly designed for reference-based evaluation.
While BARTScore and GPTScore can also be applied in a reference-free manner by directly comparing generated text to source sentences, their effectiveness is noticeably context-sensitive especially in QA evaluation due to substantial divergence between answers and questions. 

\paragraph{Behavior Consistency} 
\citet{jang2022becel} introduced the BECEL benchmark, evaluating language models across tasks like natural language inference and semantic analysis, highlighting the importance of behavioral consistency. 
\citet{asai2020logic} and \citet{ribeiro2019red} improved consistency in question-answering by enriching training data with logical and linguistic insights.
\citet{cohen2023lm} proposed the LM vs LM framework for factuality evaluation through cross-examination between LMs, using multi-turn interactions to uncover inconsistencies. However, this model relies on the subjective judgment of the examiner LM, introducing potential bias. Our approach contrasts by offering an objective mechanism for identifying inconsistencies without the constraints of LM vs LM, such as needing LMs with over 10 billion parameters and reliance on instruction-folowing and long-context reasoning ability. Additionally, unlike LM vs LM, which depends solely on consistency for correctness, our method integrates external knowledge for a more comprehensive evaluation.

\section{Conclusion}
We introduce \texttt{TrustScore}, a novel reference-free evaluation framework that assesses the trustworthiness of LLM responses.  
Our experiments, conducted on our newly curated MixedQA dataset, demonstrate 
effectiveness  
of the \texttt{TrustScore} framework.
\texttt{TrustScore} exhibits a strong correlation with human judgments, surpassing existing reference-free metrics, and achieves results comparable to reference-based metrics. As for future work, we might look into how to combine \texttt{TrustScore} with model editing methods~\citep{HLTW+2023}. 

\section*{Acknowledgments}
This work is supported by Huawei’s Dean's Funding (C-00006589) and the UKRI Centre for Doctoral Training in Natural Language Processing, funded by the UKRI (grant EP/S022481/1).

\bibliography{iclr2024_conference}
\bibliographystyle{iclr2024_conference}

\newpage
\appendix
\section{Appendix}
\label{sec:appendix}
\subsection{Limitations}
Due to the high cost of collecting human evaluation as gold labels and executing LLMs like GPT-3.5, which we include in our experiments, we do not test \texttt{TrustScore} on large-scale data. Instead, we test it on MixedQA, which consists of 1000 open-ended questions randomly collected from five question-answering datasets.

In this work, we do not test \texttt{TrustScore} on long-form questions that require elaborate and in-depth answers. We contend that our general framework can be adapted to the long-form question-answer task. We will leave it in our future work.

\subsection{Ethics Statement}
The MixedQA dataset is collected from several datasets that are public for academic use and do not contain sensitive information.  We will make the dataset and the codebase of the project freely available online for academic use.

Our human evaluation involves human participation. Two annotators were recruited and trained. They were asked to annotate the correctness of answers and filter out samples that might cause ethical problems. No sensitive personal information is involved in the process.

\subsection{Data Selection Criteria for MixedQA}
\label{app: data selection}
We collected MixedQA from a diverse set of five datasets:

\paragraph{Natural Questions  \citep{kwiatkowski2019natural}} This dataset consists of questions extracted from web queries, with each question accompanied by a corresponding Wikipedia article containing the answer.

\paragraph{WebQuestions \citep{berant2013semantic}} This dataset comprises questions originating from web queries, meticulously matched with entries in FreeBase.

\paragraph{TriviaQA \citep{joshi2017triviaqa}} This dataset incorporated a collection of questions sourced from Quiz League websites. Each question in this dataset is complemented by web pages and Wikipedia searches that may potentially contain the answer.

\paragraph{HotpotQA \citep{yang2018hotpotqa}} This dataset primarily contains Wikipedia-based multi-hop questions, with sentence-level supporting facts annotation.

\paragraph{PopQA \citep{mallen2022not}} This dataset is composed of questions aimed at covering factual information in the less frequently explored areas, essentially addressing the long tail of information.

We extracted questions exclusively from the test or development sets, discarding the supporting context to adapt to a closed-book setting. Our data collection process involved the following steps:
\begin{enumerate}
    \item Open-Ended Questions Extraction: Initially, we utilized an automated script to filter out comparison questions (e.g.,``Who was born earlier, Emma Bull or Virginia Woolf?") and ``yes or no" questions (e.g., ``Is Ferocactus a type of plant?") in order to focus on open-ended questions.
    \item Input to Language Models: We input the filtered questions, devoid of any contextual information, to three language models: FLAN-T5-XXL, LLaMA-7B, and GPT-3.5-Turbo.
    \item Answerable Questions Extraction: We observed that FLAN-T5 occasionally generated subpar answers, characterized by repetitive words or even blank responses. LLaMA-7B, on the other hand, sometimes failed to provide a relevant answer and instead repeated the original question. ChatGPT's responses, meanwhile, often fell into the category of inconclusive answers, with statements like ``I am not sure," ``I don't know," or ``The information is not provided in the question." Such responses lacked the necessary informative content required for trustworthiness verification. To address this issue, we employed an automated script to filter out questions with such responses, retaining only those questions that could be answered by the three LLMs.
    \item Random Sampling: Finally, to ensure a balanced and representative dataset, we randomly selected 200 questions from each of the five datasets for \texttt{TrustScore} evaluation.
\end{enumerate}

\subsection{Human Evaluation Details}
\label{app:ann}
The 1,000 questions are evenly distributed between two annotators.
Each question accompanied by answers from three LLMs: FLAN-T5, LLaMA, and GPT-3.5. These questions were also paired with ground truth answers from the original dataset. During our pilot study, we discovered a number of incorrect ground truth answers.
As a result, annotators were advised to use these answers as a guide and to verify the correctness of answers via manual internet searches.

\begin{table}[htbp]
\centering
\small
\begin{tabular}{cc}
\hline
System & Cohen's $\kappa$  \\ \hline
FLAN-T5 & 0.88 \\ 
LLaMA & 0.82  \\
GPT & 0.92 \\ \hline
Overall & 0.89 \\ \hline

\end{tabular}%
\caption{Inter-annotator agreement for the double annotations on fifty examples.}
\label{tab:inter-agg}
\end{table}

We evaluated the reliability of human annotations by measuring the agreement between annotators. Table~\ref{tab:inter-agg} shows the results of the Cohen’s $\kappa$ \cite{cohen1960coefficient}, which accounts for agreement that is expected by chance. The annotations on answers of FLAN-T5, LLaMA, and GPT have Cohen’s $\kappa$ values of 0.88, 0.82, and 0.92, respectively, indicating a high level of agreement. The overall Cohen’s $\kappa$ value is 0.89, confirming the quality of the annotations in the MixedQA dataset.


\subsection{Implementation Details}
\label{app:implementation}
\subsubsection{Details of the Substitution Algorithm}
\label{app:sub_alg}
To generate high-quality distractors, we propose a substitution algorithm that produces a dictionary of substitute candidates $D$, where keys represent words to be substituted and the values are the corresponding potential substitute words.
As shown in Algorithm\ref{alg: priority}, the substitution algorithm involves three functions to search for substitute candidates for a given entity or word: \textit{find\_dbpedia\_ents}, \textit{find\_candidates\_by\_embedding}, and \textit{find\_candidate\_by\_pos}. 
The \textit{find\_dbpedia\_ents} function retrieves DBpedia entities with the same DBpedia types~\citep{HBXLP2022,HBXLP2023} as the entities to be substituted. We employ the \href{https://spacy.io/universe/project/spacy-dbpedia-spotlight}{\texttt{DBpedia Spotlight for Spacy}} to detect DBpedia entities in response $r$ and obtain their corresponding DBpedia types. For example, the DBpedia types for an entity like "Kobe" might include "DBpedia:Person," "DBpedia:Athlete," and "DBpedia:BasketballPlayer." Subsequently, we use SPARQL to search for DBpedia entities that satisfy these types, starting with entities that match all types and gradually lowering the number of required matching types if necessary. The \textit{find\_candidates\_by\_embedding} function identifies the most similar words by computing cosine similarity between the mean of the projection weight vectors of the words to be substituted and the vectors for each word in a pretrained word2vec model (\href{https://github.com/RaRe-Technologies/gensim-data}{\texttt{word2vec-google-news-300}}). The \textit{find\_candidate\_by\_pos} function randomly selects words with the same POS tag from the \texttt{word2vec-google-news-300} corpus.

Once we obtain the substitute candidate dictionary $D$, we generate a set of distractors by replacing one word or entity in response $r$ with its candidates in each iteration. This approach minimizes the edit distance between the distractors and response $r$, making it more challenging for LLM $M$ to distinguish between the options.

\begin{algorithm*}
\scriptsize
    \SetKwInput{KwInput}{Input}                
    \SetKwInput{KwOutput}{Output} 
    \DontPrintSemicolon
    \KwInput{Question $q$, answer $a$, candidate minimum number $n$}
    \KwOutput{Substitute candidate dictionary $D$}
    
      \SetKwFunction{FMain}{Main}
      \SetKwFunction{Fent}{Find\_candidates\_for\_entities}
      \SetKwFunction{FWord}{Find\_candidates\_for\_words}
    
      \SetKwProg{Fn}{Function}{:}{}
      \Fn{\Fent{$entities\_list$, $num$}}{
       $candidate\_dict$ = $defaultdict(list)$ \\
       
       \tcc{search candidates for DBpedia entities (source: DBperdia)}
        \For{each $entity$ in $entities\_list$ }{
            \If{ the $entity$ is a DBpedia entity}{
                $candidate\_dict[entity]$ = find\_dbpedia\_ents$(entity, num)$
                }
        }
        
        delete the item whose value is an empty list in $candidate\_dict$ \\
        \tcc{search candidates for entities (source: embedding similarity)}
        \If{the candidate number in $candidate\_dict$ < $num$}{
            \For{each $entity$ in $entities\_list$ }{
                 $candidate\_dict[entity]$ += find\_candidates\_by\_embedding$(entity, num)$
            }
        }
        
        delete the item whose value is an empty list in $candidate\_dict$ \\
        \tcc{search candidates for each word in entities (source: embedding similarity)}
        \If{the candidate number in $candidate\_dict$ < $num$ and there exits multi-word entity in $entities\_list$}{
            \For{each $entity$ in the multi-word entities}{
                $candidate\_dict[entity]$ += [replace the non-high frequency word in the entity with the candidates searched through find\_candidates\_by\_embedding$(word, num)$]
            }
        }
        
        delete the item whose value is an empty list in $candidate\_dict$ \\
        \tcc{search candidates for entities (source: randomly selected from corpus with same pos)}
        \If{the candidate number in $candidate\_dict$ < $num$}{
           \For{each $entity$ in $entities\_list$ }{
               $candidate\_dict[entity]$ += find\_candidates\_by\_pos$(entity, num)$
           }
        }
        
            \KwRet $candidate\_dict$\;
      }
      \;
    
      \SetKwProg{Fn}{Function}{:}{}
      \Fn{\FWord{$words\_list$, $num$}}{
            $candidate\_dict$ = $defaultdict(list)$ \\
            \tcc{search candidates for words (source: embedding similarity)}
            \For{$word$ in $words\_list$}
            {
                $candidate\_dict[word]$ = find\_candidates\_by\_embedding$(word, num)$    
            }
            
            delete the item whose value is an empty list in $candidate\_dict$\\
            \tcc{search candidates for words  (source: randomly selected from corpus with same pos)}
            \If{the candidate number in $candidate\_dict$ < $num$}{
            \For{$ent$ in $words\_list$}{
                $candidate\_dict[word]$ = find\_candidates\_by\_pos$(word, num)$
            }
            }
            \KwRet  $candidate\_dict$\;
      }
      \;
    
      \SetKwProg{Fn}{Function}{:}{\KwRet}
      \Fn{\FMain}{
            $ents_1, ents_2$ = get entities in answer $a$ \tcp*{1 only in answer, 2 also in question}
            $nncds_1, vbjjs_1, nncds_2, others$ = split words in answer $a$ with their pos tag
            
            $level_1, level_2, level_3, level_4 $ = $ ents_1, nncds_1, vbjjs_1 + ents_2 + nncds_2, others$ \\
            \;
            $D$ = $defaultdict(list)$ \\
    
            \If{$level_1$}
                { $D$ = Find\_candidates\_for\_entities$(level_1, n)$
                }
            \ElseIf{$level_2$}
                {
                $D $= Find\_candidates\_for\_words$(level_2, n)$
                }
            \ElseIf{$level_3$}{
                \If{$vbjjs_1$}{
                 $D$ = Find\_candidates\_for\_words$(vbjjs_1, n)$
                }
            
                \If{the candidate number in $D$ < $n$ and $ents_2$ exits}{
                    $D$ = merge($D$, Find\_candidates\_for\_entities$(ents_2, n)$)
                }
            
                \If{the candidate number in $D$ < $n$ and $nncds_2$ exits}{
                    $D$ = merge($D$, Find\_candidates\_for\_words$(nncds_2, n)$)
                }
            }
            \ElseIf{$level_4$}{
                $D$=Find\_candidates\_for\_words$(level_4, n)$
            }
            \Else{
            print("I did not find the token that needs to be replaced.")
            }
            \KwRet D\;
      }
\caption{Priority-based Substitution Algorithm}
\label{alg: priority}
\end{algorithm*}

\newpage
\subsubsection{Trust$_{FC}$}
\label{app:fce} 
We retrieve the top 10 related passages as evidence $e$ for the query \textit{"answer $ar$" correctly answers the question of "question $q$"} using a BM25-based \cite{robertson2009probabilistic} method from the \texttt{wikipedia-dpr} corpus~\cite{karpukhin2020dense}. Following \citet{qin2023chatgpt}, we employ \href{https://platform.openai.com/docs/models/gpt-3-5}{\texttt{GPT-Turbo-3.5}} as an entailment model to judge whether the evidence $e$ can validate the response $r$ generated for question $q$.

\subsubsection{Scoring Criterion}
\label{app:score} 
The scoring criterion is shown in Table~\ref{tab:score}.

\begin{table}[h]
\caption{The scoring criterion for Trust$_{OV}$ which integrate the Trust$_{BC}$ and Trust$_{FC}$.}
\centering

\begin{tabular}{ccc}
\hline
\begin{tabular}[c]{@{}c@{}}\texttt{Trust$_{BC}$}\end{tabular} & \begin{tabular}[c]{@{}c@{}}\texttt{Trust$_{FC}$}\end{tabular} & Trust$_{OV}$ \\ \hline
Consistent   & Support  & 1 \\
Inconsistent & Support  & 0.8 \\
Consistent  & Neutral & 0.6 \\
Inconsistent & Neutral & 0.4 \\
Consistent   & Contradict & 0.2 \\
Inconsistent & Contradict & 0 \\ \hline
\end{tabular}%
\label{tab:score}
\end{table}

\subsubsection{Hyper-Parameters}
\label{app:para} 
FLAN-T5 and LLaMA employ a greedy decoding strategy with a maximum length of 1042 tokens.
GPT-3.5 uses a temperature setting of 0 with a maximum length of 1042 tokens.
The maximum check limit of behavior consistency evaluator is set as 10.

\subsubsection{Prompts}
\label{app:prompt} 
Table~\ref{tab:prompt_2} presents the prompt used in closed-book question answering task.

\begin{table*}[h]
\caption{The prompt for Close-book Question Answering}
    \centering
    \small
    \begin{tabular}{p{.95\textwidth}}       
    \hline
    \textbf{Prompt for Close-book Question Answering} \\
    \hline
        INSTRUCTION: Please give answers to the following questions about knowledge.\\
        
        Question: who has been ranked no. 1 in the latest football rankings announced by fifa? \\
        Answer: Germany has been ranked no. 1 in the latest football rankings announced by fifa. \\
        \\
        
        Question: who sings i just want to use your love tonight? \\
        Answer: English rock band the Outfield sings i just want to use your love tonight. \\
        \\
        
        Question: where was the movie the glass castle filmed? \\
        Answer: The movie the glass castle was filmed in Welch, West Virginia.\\
        \\
        
        Question: who was the first lady nominated member of the rajya sabha? \\
        Answer: Mary Kom was the first lady nominated member of the rajya sabha.\\
        \\
        
        Question: what is the tiger's name in life of pi? \\
        Answer: Richard Parker is the tiger's name in life of pi.\\
        \\
        
        Question: \{Q\} \\
        Answer: \\
    \hline
    \end{tabular}
    \label{tab:prompt_2}
\end{table*}

\newpage
Table~\ref{tab:prompt_1} presents the prompt used in behavior consistency evaluator.
\begin{table*}[h]
\caption{The prompt for Behavior Consistency Evaluation}
    \centering
    \small
    \begin{tabular}{p{.95\textwidth}}      
    \hline
    \textbf{Prompt for Behavior Consistency Evaluation} \\
    \hline
        INSTRUCTION: Please give answers to the following multi-choice questions about knowledge. \\

        Question: who has been ranked no. 1 in the latest football rankings announced by fifa? \\
        A) Germany has been ranked no. 1 in the latest football rankings announced by fifa. \\
        B) India has been ranked no. 1 in the latest football rankings announced by fifa. \\
        C) Canada has been ranked no. 1 in the latest football rankings announced by fifa. \\
        D) Austria has been ranked no. 1 in the latest football rankings announced by fifa. \\
        E) None of above. \\
        Answer: E \\
        \\
        
        Question: who sings i just want to use your love tonight? \\
        A) Latin rock band the Outfield sings i just want to use your love tonight. \\
        B) English Power pop band the Outfield sings i just want to use your love tonight. \\
        C) English rock band the Outfield sings i just want to use your love tonight. \\
        D) English melodic sensibility band the Outfield sings i just want to use your love tonight. \\
        E) None of above. \\
        Answer: C \\
        \\
        
        Question: where was the movie the glass castle filmed? \\
        A) The movie the glass castle was filmed in London. \\
        B) The movie the glass castle was filmed in Welch, West Virginia. \\
        C) The movie the glass castle was filmed in Philadelphia. \\
        D) The movie the glass castle was filmed in Budapest. \\
        E) None of above. \\
        Answer: B \\
        \\
        
        Question: who was the first lady nominated member of the rajya sabha? \\
        A) William Randolph Hearst was the first lady nominated member of the rajya sabha. \\
        B) Jesse Speight was the first lady nominated member of the rajya sabha. \\
        C) Thurlow Weed was the first lady nominated member of the rajya sabha. \\
        D) Mary Kom was the first lady nominated member of the rajya sabha. \\
        E) None of above. \\
        Answer: D \\
        \\
        
        Question: what is on a mcchicken sandwich from mcdonalds? \\
        A) A breaded chicken patty is on a mcchicken sandwich from mcdonalds. \\
        B) A Hot dog chicken patty is on a mcchicken sandwich from mcdonalds. \\
        C) A breaded Bacon is on a mcchicken sandwich from mcdonalds. \\
        D) A breaded Teriyaki chicken is on a mcchicken sandwich from mcdonalds. \\
        E) None of above. \\
        Answer: A \\
        \\
        
        Question: \{Q\} \\
        Answer: \\
    \hline
    \end{tabular}
    \label{tab:prompt_1}
\end{table*}

Table~\ref{tab:factual} presents the prompt used in factual consistency evaluator.

\begin{table*}[h]
\caption{Prompt for Fact Checking}
    \centering
    \small
    \begin{tabular}{p{.95\textwidth}}    
    \hline
    \textbf{Prompt for Fact Checking}\\
    \hline
    INSTRUCTION: Assess whether the provided evidence aligns with or contradicts the given question-answer pair, and categorize the relationship as either `support', `contradict', or `neutral'.\\
    Evidence: [evidence $e$]\\
    Question: [question $q$]\\
    Answer: [answer $a$]\\
    Relation:\\

    \hline
    \end{tabular}
    \label{tab:factual}
\end{table*}

\subsection{Examples of Multi-Choice Questions generated for Behavior Consistency Evaluation}
\label{app:mc-q}

Table~\ref{tab:mc example} presents the examples of the multi-choice questions used in Behavior Consistency Evaluator

\begin{table*}[h]
\caption{Examples of Multi-Choice Questions Generated for Behavior Consistency Evaluation.}
    \centering
    \small
    \begin{tabular}{p{.95\linewidth}}    
    \hline
    \textbf{Examples of Multi-Choice Questions Generated for Behavior Consistency Evaluation}\\
    \hline
    When did all night long come out lionel richie?\\
    A) All night long came out in 1975. \textcolor{red}{[distractor]}\\
    B) All night long came out in 1986. \textcolor{red}{[distractor]}\\
    C) All night long came out in 1983. \textcolor{green}{[original response]}\\ 
    D) All night long came out in 1999. \textcolor{red}{[distractor]}\\
    E) None of the above.\\
    \hline

    Who built the first temple for god in jerusalem?\\
    A) Gopala III built the first temple for god in jerusalem. \textcolor{red}{[distractor]}\\
    B) King Solomon built the first temple for god in jerusalem. \textcolor{green}{[original response]}\\
    C) Mohammad Hatta built the first temple for god in jerusalem. \textcolor{red}{[distractor]}\\
    D) Alexander the Great built the first temple for god in jerusalem. \textcolor{red}{[distractor]}\\
    E) None of the above.\\
    \hline
    which state was returned to spain after the revolutionary war?\\
    A) Illinois was returned to spain after the revolutionary war. \textcolor{red}{[distractor]}\\
    B) Louisiana was returned to spain after the revolutionary war. \textcolor{red}{[distractor]}\\
    C) Florida was returned to spain after the revolutionary war. \textcolor{green}{[original response]}\\
    D) Lincolnshire was returned to spain after the revolutionary war. \textcolor{red}{[distractor]} \\
    E) None of the above. \\
     \hline

    \end{tabular}
    \label{tab:mc example}
\end{table*}

\end{document}